%% file: main.tex
\documentclass[conference]{IEEEtran}
\IEEEoverridecommandlockouts

\usepackage{cite}
\usepackage{amsmath,amssymb,amsfonts}
\usepackage{algorithmic}
\usepackage{graphicx}
\usepackage{textcomp}
\usepackage{url} 
\usepackage{xcolor}
\newcommand{\ie}{{\itshape i.e.}, }

\def\BibTeX{{\rm B\kern-.05em{\sc i\kern-.025em b}\kern-.08em
    T\kern-.1667em\lower.7ex\hbox{E}\kern-.125emX}}
\begin{document}

\title{Orchestrating Agents and Data for Enterprise:\\ A Blueprint Architecture for Compound AI}

\author{
\IEEEauthorblockN{Eser Kandogan, Nikita Bhutani, Dan Zhang, Rafael Li Chen, Sairam Gurajada, Estevam Hruschka}
\IEEEauthorblockA{\textit{Megagon Labs, USA} \\
Mountain View, CA, U.S.A \\
\{eser, nikita, dan\_z, rafael, sairam, estevam\}@megagon.ai}
}

\maketitle

\begin{abstract}
Large language models (LLMs) have gained significant interest in industry due to their impressive capabilities across a wide range of tasks. 
However, the widespread adoption of LLMs presents several challenges, such as integration into existing applications and infrastructure, utilization of company proprietary data, models, and APIs, and meeting cost, quality, responsiveness, and other requirements. To address these challenges, there is a notable shift from monolithic models to compound AI systems,  with the premise of more powerful, versatile, and reliable applications. However, progress thus far has been piecemeal, with proposals for agentic workflows, programming models, and extended LLM capabilities, without a clear vision of an overall architecture.
In this paper, we propose a `blueprint architecture' for compound AI systems for orchestrating agents and data for enterprise applications. In our proposed architecture the key orchestration concept is `streams' to coordinate the flow of data and instructions among agents. Existing proprietary models and APIs in the enterprise are mapped to `agents', defined in an `agent registry' that serves agent metadata and learned representations for search and planning. Agents can utilize proprietary data through a `data registry' that similarly registers enterprise data of various modalities. Tying it all together, data and task `planners' break down, map, and optimize tasks and queries for given quality of service (QoS) requirements such as cost, accuracy, and latency. We  illustrate an implementation of the architecture for a use-case in the HR domain and discuss opportunities and challenges for `agentic AI' in the enterprise.
\end{abstract}


\input{intro}

\input{example}

\input{related}

\input{blueprint}

\input{casestudy}
\input{discussion}
\input{conclusion}

\bibliographystyle{plain} 
\bibliography{paper}

\end{document}

%% file: intro.tex
\section{Introduction}
\label{sec:intro}

Despite their impressive performance across a wide range of tasks, adopting LLMs in enterprise applications is challenging \cite{wu2024exploring,kasneci2023chatgpt, wang2023gemini, liang2021towards, xi2023rise}. On top of well-known issues with LLMs, such as sensitivity \cite{lu2021fantastically}, hallucination \cite{semnani2023wikichat, ji2023survey}, fairness \cite{li2023survey}, and ethics \cite{thoppilan2022lamda},
enterprise companies want to exploit their proprietary data, models and APIs to continue their strategic advantage over their competitors. However, integration with LLMs presents additional challenges in terms of cost, control, privacy, and quality.

To mitigate these challenges, there is broad recognition in the industry to approach this as a `systems' problem, \ie integrate LLMs 
\emph{compound AI systems} along with other components such as tools and retrievers \cite{topsakal2023creating,schick2024toolformer,compound-ai-blog,zhu2024knowagent}. Relevant work includes: 

\begin{itemize}
  \item \emph{new  components}, such as retrieval-augmented generation models ~\cite{lewis2020retrieval}, automatic fact verifiers~\cite{semnani2023wikichat, li2023self}, content moderation modules~\cite{yuan2024rigorllm},
  \item \emph{new workflow patterns} such as chain-of-thought reasoning~\cite{wei2022chain}, mixture-of-experts~\cite{du2022glam} and others \cite{paranjape2023art, zhang2023ecoassistant, wang2024chain}, to improve reasoning, accuracy, 
  \item \emph{new LLM capabilities} such as tool calling ~\cite{schick2024toolformer,thoppilan2022lamda, hao2024toolkengpt, liang2024taskmatrix, wang2024tools}, tool making \cite{cai2023large}, integrating  reasoners \cite{lewkowycz2022solving} among others, and 
  \item \emph{new programming models and frameworks} such as DSPy, to program and optimize agentic workflows ~\cite{khattab2023dspy,10.1145/3586183.3606763, crawford2024bmw, wu2023autogen, Liu_LlamaIndex_2022, liu2024agentlite, lu2024chameleon, Xie2023openagents,chen2023autoagents,zhou2023agents,chen2023agentverse, liu2024declarative}.
\end{itemize}

\noindent Despite progress a unifying architecture is still yet to emerge, with several questions remaining unanswered. How to address the large scale of data and services typical in the enterprise? How to support a large number of applications with varying workloads yet be able to optimize overall? How to enable agents to exploit data of various modalities, address discoverability, utility, and usability effectively? How can the architecture account for and optimize the quality of service in terms of cost, response time, and accuracy? How to enforce sufficient control in outcomes yet exploit the language understanding and knowledge in LLMs?

We envision a blueprint architecture, where LLMs still play an important role but are not necessarily the 'be-all and end-all'. Unlike previous proposals, where LLMs handle everything - from planning, to invoking tools and fetching data - our architecture puts LLMs within the confines of a larger architecture with dedicated components for specific functions with designed logic and controls. Our proposed architecture includes a multitude of components such as agents, registries, and planners, among others, to plan and break down complex tasks, discover and query proprietary data, and exploit proprietary models and services. All these components seamlessly work together with the support of a streaming database and an architecture orchestrating the flow of data and instructions, deployed in a distributed system with containers running each component, configured to scale and restart on failure.

Key factors for the design of the architecture include:
\begin{itemize}
\item \emph{seamless integration} into existing enterprise infrastructure through appropriate touch points and interfaces, facilitating extensibility, customizability, and reusability,

\item explicit representation of the \emph{orchestration} of data and control among components through streams in databases, providing observability and full controllability,

\item \emph {coordination} of tasks through declarative plans and ad hoc planners to flexibly drive a variety of use-cases, ensuring transparency while laying the foundation for optimization, and

\item \emph {optimization} of the entire system, from task execution to data retrieval, while adhering to quality of service (QoS) constraints to maximize utility and accountability.

\end{itemize}
   
As such, we go beyond existing work~\cite{wu2023autogen, Liu_LlamaIndex_2022} with specific support for integration, orchestration, coordination, and optimization, while also drawing inspirations from database and systems literature to address above challenges. More specifically, existing multi-agent frameworks lack the necessary abstractions for  orchestration and coordination, rely heavily on textual data flowing between agents, support limited optimization objectives primarily on LLMs, and lack mechanisms of discovery and representation of existing data, models, and agents. Unlike other frameworks, we squarely focus on the key abstractions in the framework, to standardize the flow of data and controls among agents, to represent existing models and data through registries, and to plan and optimize supporting multiple objectives.

The remainder of the paper is organized as follows. We first introduce an example enterprise company and a couple of scenarios to motivate the design of the architecture. Next, we introduce our proposed blueprint architecture and describe the key components in detail. Then, we present a specific implementation of the architecture for an HR use-case. Finally, we discuss open research problems and argue for a systems approach incorporating ideas from artificial intelligence, database, and human-computer interaction fields.

%% file: example.tex
\section{Example: Human Resources Company}
\label{sec:example}

To facilitate discussion on design considerations and introduce our proposed architecture, we outline a typical enterprise scenario. While we present an HR scenario, the proposed architecture is not specific to any industry but rather to enterprise setting, supporting large number of services, databases, and models supporting a wide variety of applications.

Consider YourJourney, a mid-size HR company, specializing in engineering jobs, serving about 1M job seekers,  hiring managers, and recruiters for mid to large-size companies. YourJourney offers a wide range of services: job seekers can upload resumes, search and apply for jobs, while employers can post job openings, search for and match candidates, and track applications. YourJourney uses cloud providers for compute and storage infrastructure for many of these applications. Over the course of several years,  YourJourney has collected extensive resume, job posting, and application data hosted on several databases (document, relational), and trained models, including finetuned LLMs, for various tasks such as skill extraction, matching, and ranking. 

Given recent advances, the company aims to integrate LLMs to enhance existing applications (e.g., matching, search), develop new ones (e.g., conversational career advice, agentic job interview), and expand into new sectors (e.g. medical). 

While enthusiastic about new LLM capabilities, the company is also mindful of the limitations such as hallucination, bias, reliability, latency, and costs. They are considering developing modules for content moderation and explanation and investigating retrieval-augmentation to utilize their data.

Below we describe a few application scenarios the company is planning to build, along with a running example to use throughout the paper.

\subsection{Scenario I: Career Assistance.}
\label{sec:example_career}
One novel application YourJourney is considering is to build a conversational career assistant to support job seekers in exploring companies and roles, conducting job searches, and supporting their careers, to respond to inquiries such as:
``\textit{Do large corporations require extensive sales experience?}'',
``\textit{What data scientist positions does Google have in Mountain view?}'',
``\textit{I want to be a data scientist... what are the required skills?}''

\textbf{Running Example}: We will use the following  query as the running example throughout the paper: ``\textit{I am looking for a data scientist position in SF bay area.}'' 

\subsection{Scenario II: Agentic Employer}
\label{sec:example_agentic_employer}
Agentic Employer application allows employers to sift through applicants to their job posts. In a conversational experience, employers can utilize sophisticated predictive models to rank and cluster candidates, ask questions about the applicants, create lists interactively by add and remove applicants through queries and generate summaries. In many of these the application uses LLMs as well as predictive models built with data from relational and document databases.

%% file: related.tex
\section{Related Work}
\label{sec:related}

Below, we discuss work related to the design of compound AI systems that have emerged in both industry and research.

\subsection{Augmenting Models with Tools and Data}

An emerging trend is to extend LLMs capabilities with external tools and data to perform  tasks and access relevant knowledge~\cite{cmdbench,kandogan2024blueprint} by conditioning generation with retrieval to improve accuracy and relevance. Approaches range from utilizing a single data source  \cite{lewis2020retrieval} to heterogeneous, multi-modal data sources \cite{chen2023symphony}, with the execution being a single-step \cite{ram2023context,yoran2023making} or interleaved with reasoning \cite{yao2022react}. 

Other approaches extend LLMs to include verification~\cite{zhao2023verify}, summarization~\cite{wu2309hayate}, explanation~\cite{mishra2023characterizing}, and self-reflection modules~\cite{asai2023self,maekawa2024retrieval}. Verification modules validate content against trusted sources for accuracy. Summarization modules condense information into concise and coherent summaries. Explanation modules aim to provide detailed insights and enhance transparency. Self-reflection modules enable the model to assess outputs for coherence, consistency, and correctness enhancing overall quality.

All these approaches attest to the importance of extending LLMs with external data and tools, yet they often lack an abstraction and overarching architecture for proprietary models, data, and APIs to drive multitude of use-cases. Such abstractions are crucial in an enterprise environment, where different applications may require varying combinations of these modules.

\subsection{Agentic Workflows, Planning, and Orchestration}

Enterprise architecture needs to support a variety of use-cases and applications requiring different workflows, data sources, and models. For example, a matching application can use a retrieve-generate-explain pipeline using application data.  In agentic workflows, LLMs reason and plan, generating reasoning steps and task-specific actions\cite{wei2022chain,yao2022react}, which are then delegated to specialized agents and APIs. To decide how to delegate tasks to tools and APIs, models are often trained with few-shot or self-supervised learning methods \cite{cai2023large,schick2024toolformer, hao2024toolkengpt,lu2024chameleon}. In the mean time, frameworks are being develop to optimize multiple calls to models and other components for diverse workloads \cite{topsakal2023creating,Liu_LlamaIndex_2022,wu2023autogen}.

Across these approaches, the control of execution varies between static and dynamic approaches. In static execution, a central LLM plans how to execute a task with different agents and tools, selects the optimal plan, and executes it. Conversely, in dynamic execution, the next step is determined dynamically based on the output of the previous step without a pre-defiend execution strategy. When the execution is centralized, the LLM serves as the core, managing how tasks are dispatched to tools, handling responses and planning subsequent steps. In contrast, in a decentralized execution, each agent/tool within the system can independently decide which agent to invoke next or when to terminate execution. 

In an enterprise, different applications may require diverse workflows, planning, and orchestration strategies. Currently, there is no unified architecture that supports dynamic tool chaining along with various styles of planning—whether pre-wired, ad hoc or anything in between. Also, regarding orchestration observability, there are no established abstractions and repositories that represent and persist the flow data and control among agents and other components in the architecture. 

\subsection{Optimizing Workflows}

In an enterprise, the workflow for an application must be optimized to maximize the quality of responses while meeting the constraints on budget and latency. This requires collectively optimizing multiple modules in the workflow to work well together. Frameworks such as DSPy \cite{khattab2023dspy} and FrugalGPT \cite{chen2023frugalgpt} provide a general optimizer for workflows to minimize costs and improve accuracy. However, they largely focus on LLM-based tools/agents. As such, optimizing workflows in a compound AI system for a range of agents is still an open problem.

%% file: blueprint.tex
\section{Blueprint Architecture: Overview}
\label{sec:overview}

\begin{figure*}[!htb] 
  \centering
  \includegraphics[width=0.75\textwidth]{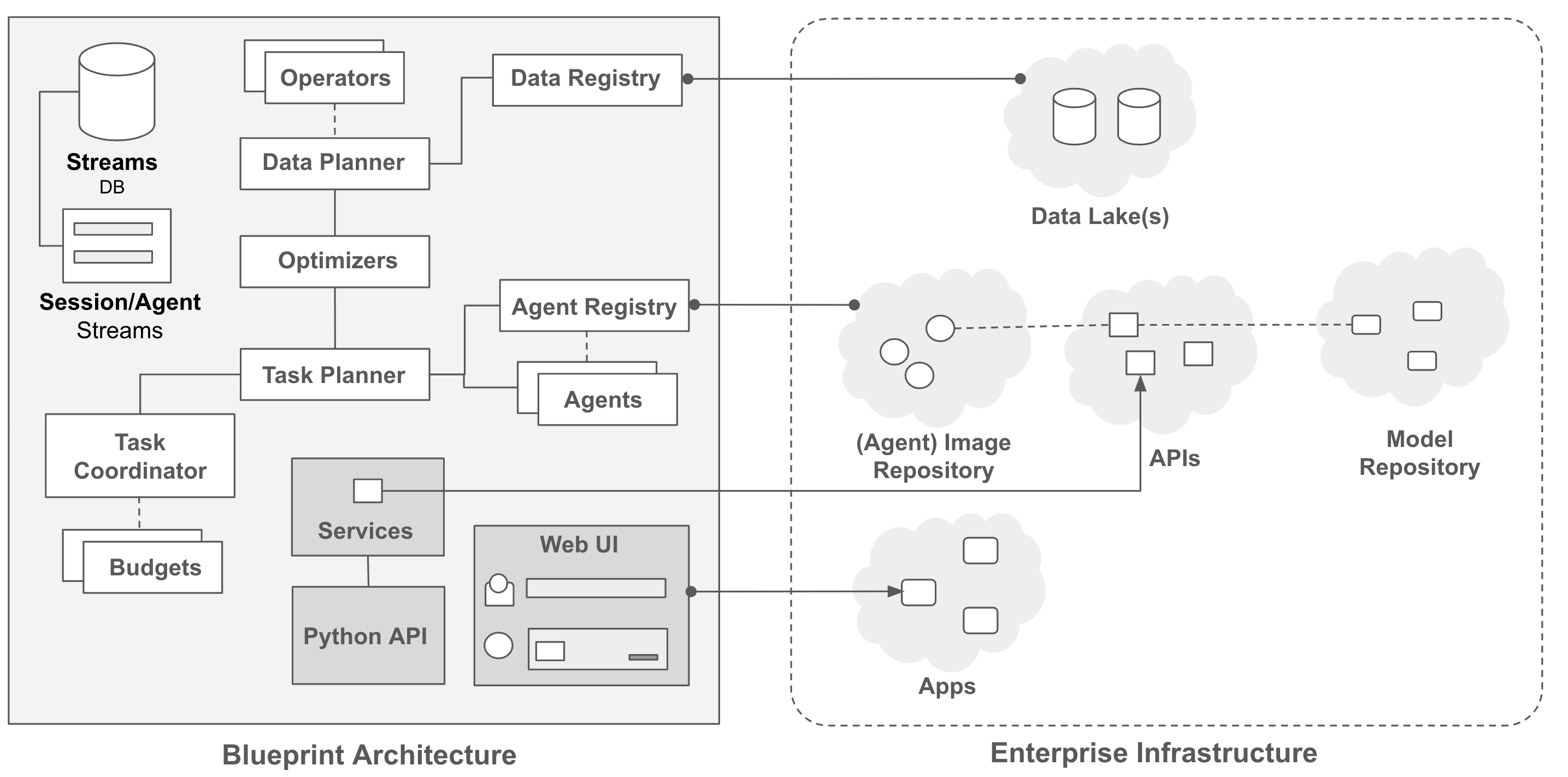}
  \vspace{-10pt}
  \caption{Blueprint Architecture: Data and Agent Registries are touch points that define existing data, models, and APIs, and services in the enterprise.}
  \label{fig:architecture} 
  \vspace{-15pt}
\end{figure*}

We now provide an overview of our proposed blueprint architecture for a compound AI system (Figure~\ref{fig:architecture}). Recall the key design goals: integration, orchestration, coordination, and optimization. We address them through the following components along with their key objectives: 

\begin{itemize}
\item \textbf{agent(s)}: maps existing models (incl. LLMs) and APIs and interfaces other agents
\item \textbf{agent registry}: serves as a metadata store for agents, and facilitates search and planning over agents using metadata
\item \textbf{task planner}: given a user utterance, create a agentic workflow, utilizing agents in the registry
\item \textbf{task coordinator}: coordinates and monitors execution of an agentic workflow, keeping track of the budget
\item \textbf{budget}: records of the current and projected QoS stats to guide execution, planning
\item \textbf{data registry}: serves as metadata store for data sources, supports operations like discovery, search and query
\item \textbf{data planner}: given a natural language query, generates a query plan, leveraging metadata from the data registry
\item \textbf{optimizer}: performs multi-objective optimization over task and data plans
\item \textbf{streams}: facilitate the flow of data and control among agents and other components
\item \textbf{session}: provides context for various agents to conduct work, captures agent activity and outputs produced
\end{itemize}

In production setting, these components are distributed across different clusters with varying compute and networking configurations. For instance, agents are deployed according to their requirements (e.g. CPU or GPU). Services that agents call to fetch data or access model predictions can similarly be deployed to clusters based on their needs. Streams database manages the flow of data and control messages among these components across different instance and itself requires substantial storage capacity and a low-latency network.

\begin{figure}[!htb] 
  \centering
  \includegraphics[width=0.8\linewidth]{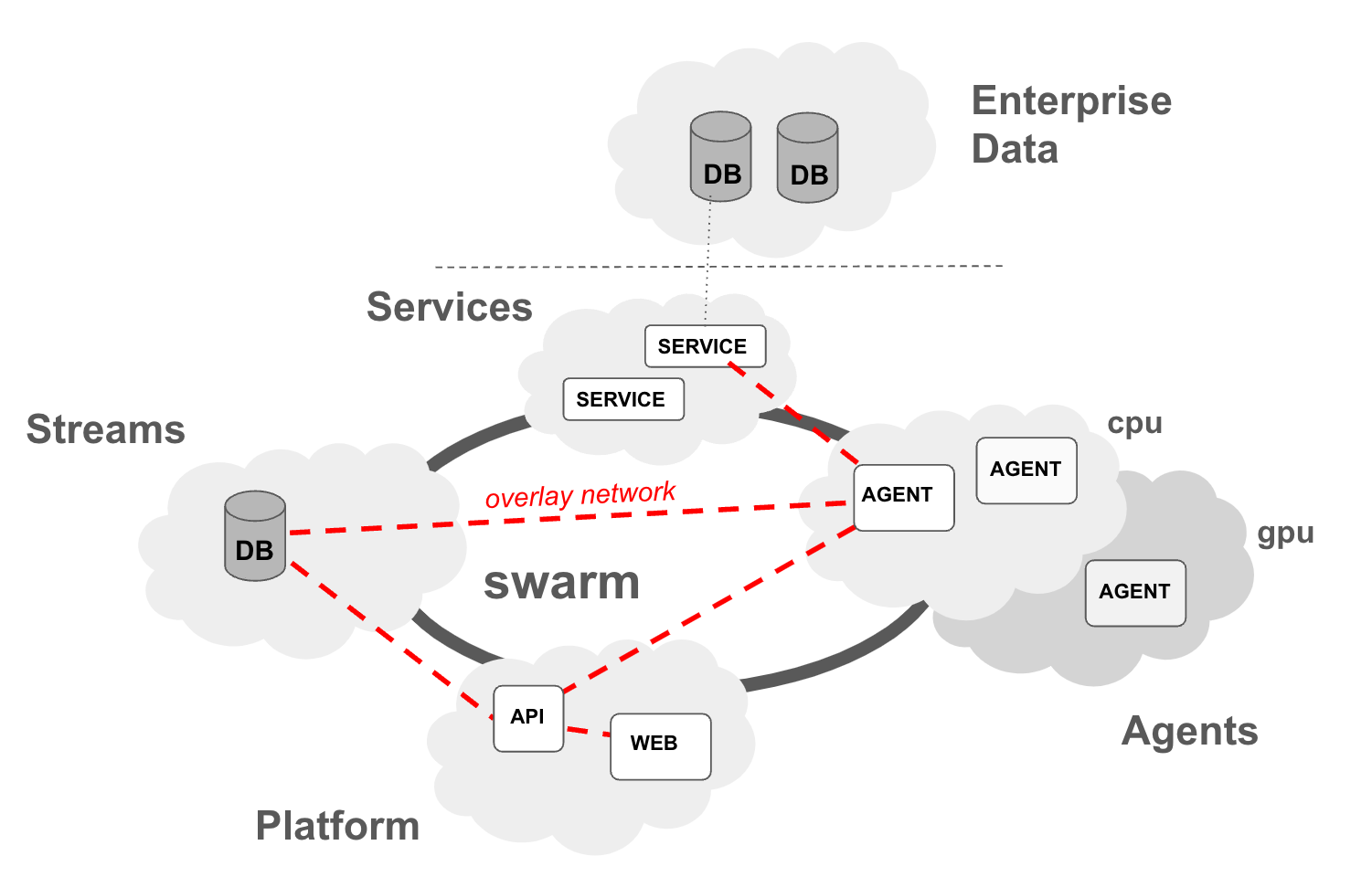}
  \caption{Deployment of blueprint architecture components in an enterprise cluster setting.}
  \label{fig:agent} 
  \vspace{-10pt}
\end{figure}

\section{Blueprint Components: Details}

We now detail each component and explain how they fulfill the design requirements for seamless integration, externalized orchestration, declarative coordination, and optimization. 

\subsection{Streams: Orchestrating Data and Control}

In the blueprint architecture, a `stream' is the central `orchestration' concept. A stream is essentially a sequence of messages, containing data or instructions, that can be dynamically produced, distributed, monitored, and consumed. For instance, user text in a chat can be modeled as a stream, with each word as an individual message. Similarly, content generated by an LLM forms another stream. Streams can contain both data and control messages: data messages facilitate data sharing between different components, while control messages allow specific instructions (e.g., invoke SQL agent) to be exchanged among components.  Components subscribe to streams to listen and receive notifications for data and control messages, and consume them to conduct work.

Streams are modeled as data structures, elevating them to first-class data resources within the system. This allows us to explicitly represent data and control exchanges,  enhancing observability and support for a variety of workflows such as fixed, open-ended, ad hoc, and pub/sub models. Being a shared data resource in the system, streams enable seamless connectivity and communication among components, even across diverse compute environments.

\subsection{Agents: Integrating Enterprise API, Models}

Agents, agent registry, and data registry serve as critical touch points and interfaces to facilitate seamless integration with existing deployed models, APIs, databases, and services in the enterprise. An agent, broadly defined as a `compute' entity responsible for executing tasks (Figure \ref{fig:agent}), extends beyond traditional definitions in the literature \cite{xi2023rise}. It encompasses any computational entity that processes input data and generates output. For instance, an LLM-based model, a task-specific CRF model, a search interface, or any API can serve as an agent in the architecture. In the context of our example from Section \ref{sec:example}, there can be an agent \textsc{Profiler}
that presents a user profile UI form to collect information from the user, and another agent \textsc{Job Matcher} to assess the match quality between a user profile and a specific job.

\begin{figure}[!htb] 
  \centering
  \includegraphics[width=\linewidth]{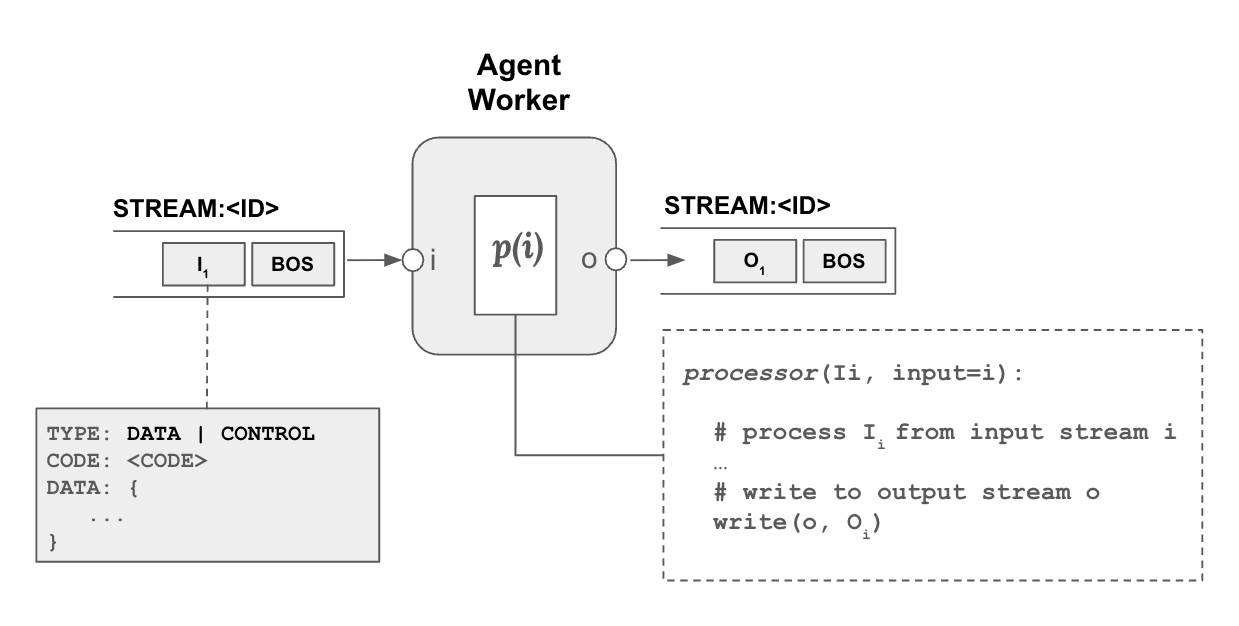}
  \caption{Agents: Triggered by data/control messages from incoming streams, agents process and produce output data and controls to output streams.}
  \label{fig:agent} 
  \vspace{-10pt}
\end{figure}

Formally, each agent is structured with input and output parameters, alongside properties that dictate its behavior. Agents utilize a \textit{processor()} function to handle incoming data and instructions, produce outputs and write into streams (Figure \ref{fig:agent}). Agents can be activated centrally through explicit instructions or in a decentralized manner by monitoring designated tags within streams, defined by inclusion and exclusion rules. For example, a message tagged \textsc{SQL} can trigger SQLExecutor agent to execute the query in the message. 

\begin{figure}[!htb] 
  \vspace{-10pt}
  \centering
  \includegraphics[width=\linewidth]{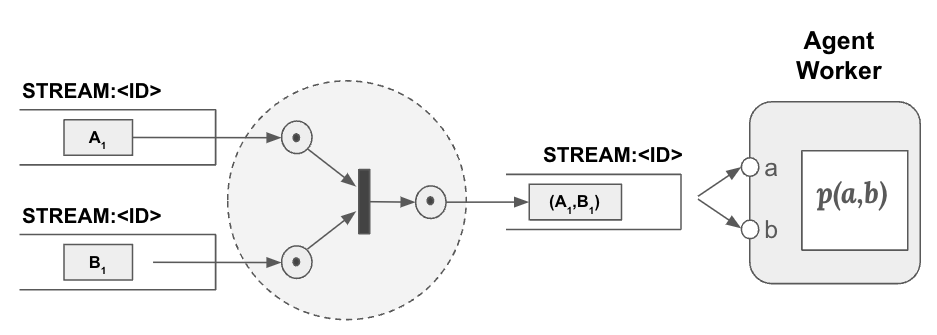}
  \vspace{-10pt}
  \caption{Agents can process data from multiple streams through a triggering mechanism inspired by PetriNets.}
  \label{fig:agents-petrinet} 
  \vspace{-10pt}
\end{figure}

When an agent is triggered, its \textit{processor()} function processes input data from one or more streams. Agents may have multiple input and output parameters, requiring preparation of all necessary input data before invoking the \textit{processor()} function. We draw inspiration from PetriNets \cite{peterson1977petri} to accomplish this (Figure \ref{fig:agents-petrinet}). We consider each input stream as a `place' holding one or more tokens (input data). Transitions occur when all places contain at least a token, allowing formation of a tuple with all input data for the processor function. Agent properties can define various configurations for triggering, such as pairing tokens from multiple streams.

Agents are deployed in containers tailored to their deployment needs, where container runs an AgentFactory server, which spawns instances of agents. When triggered, an agent can further spawn a worker, running on its own thread, while the agent continues to listen to other potential streams. This can help improve utilization and facilitate concurrency, as each agent has a pool of workers.

Agent outputs can appear in a web interface for users, though they may also just produce intermediate results that are not displayed. Simple data types (e.g., strings) in streams use straightforward renderers, while complex data like JSON employs interactive renderers enabling browsing. Agents can also generate UI forms, for example to collect user profiles, specified declaratively and displayed using UI renderers. Details on rendering and event processing via event streams are omitted here for brevity.

\subsection{Agent Registry - Mapping APIs, models}

An agent registry stores and organizes metadata about agents, including their descriptions, input and output parameters, stream inclusion/exclusion rules, deployment information such as docker images, deployment configurations, along with other properties. Agent registry facilitates search and query for efficient management and retrieval of agent metadata. 

In enterprises, existing APIs, tools, and models can be designated as agents within a registry. This registry plays a crucial role in mapping enterprise assets and making them available for integration into a comprehensive AI system. For instance, a predictive model like a job matching algorithm can be registered as an agent. It would be assigned a name (e.g., \textsc{Job Matcher}), a description, input parameters such as \textsc{Job Seeker Data}, \textsc{Jobs} and optionally \textsc{Criteria} for additional conditions. The agent can also specify an output parameter, \textsc{Matches}, with detailed descriptions including type and defaults where applicable.

Agent registry can be queried and searched by any agent. Searches can utilize keywords or vector-based techniques using learned representations derived from metadata and logs. Historical usage data can also be leveraged to compute enhanced embeddings, optimizing the search capabilities of the registry. Moreover, developers can interact with the registry through a web interface. They can register new agents, update metadata, derive new agents from existing ones, and browse and search the registry. This interface provides comprehensive functionality for managing agents within the ecosystem.

\subsection{Data Registry - Mapping Enterprise Data}

\begin{figure}[!htb] 
  \centering
  \includegraphics[width=\linewidth]{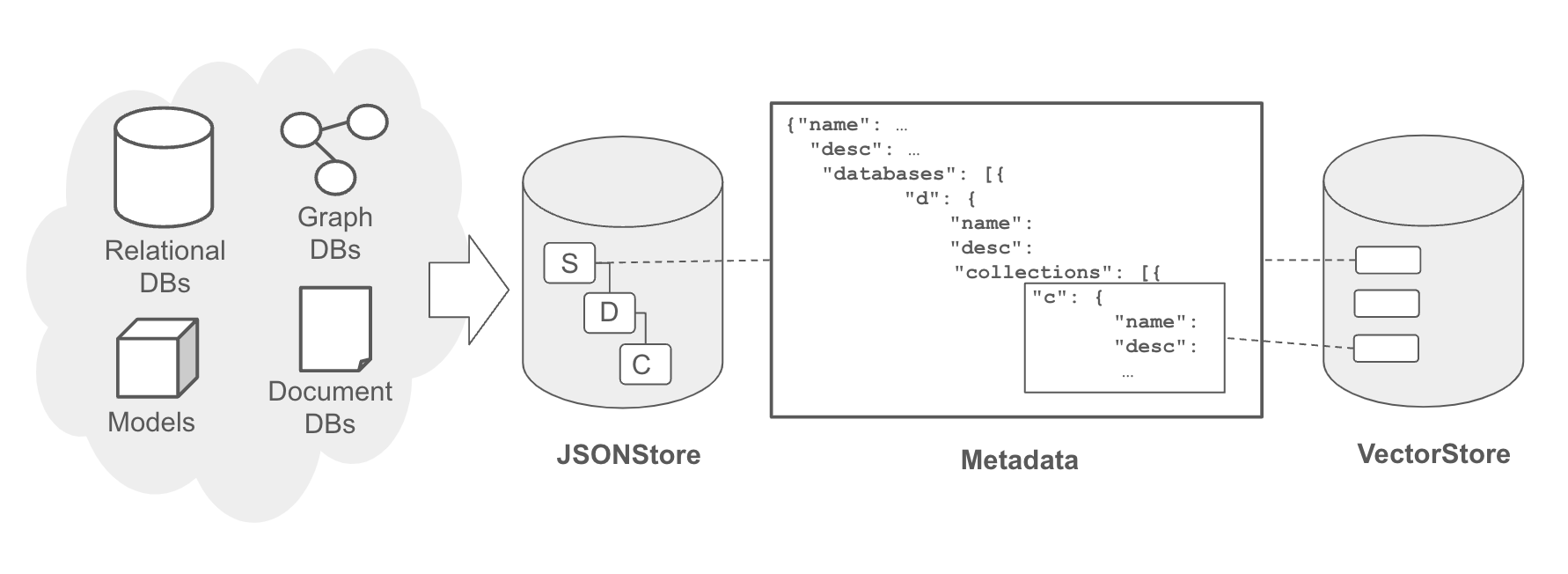}
  \vspace{-10pt}
  \caption{Data Registry}
  \label{fig:data-registry} 
  \vspace{-10pt}
\end{figure}

Analogous to an agent registry, data registry stores metadata of data in the enterprise, and as such is a key touchpoint to the existing data infrastructure (Figure \ref{fig:data-registry}). A data registry plays a crucial role in facilitating search and discovery of multi-modal enterprise data across different levels such as lakehouse, lake, source system, database, and table. It encompasses various data types including documents, relational databases, graph databases, and key-value stores. The registry helps in efficiently locating and accessing diverse data resources within the enterprise ecosystem. For instance, in our scenario, the registry could include the table holding job data, labeled as \textsc{Jobs}, detailing its descriptions and schema within the \textsc{HR} database. Another entry, \textsc{Profiles}, might represent job seeker profiles stored in a document collection, complete with associated metadata.

Metadata within a data registry typically includes names, descriptions, and details about data at various levels of granularity (e.g., tables within a database), encompassing data schema and db connection specifics. The registry also incorporates vector metadata, which captures embeddings derived from learned representations of metadata (e.g., schema details), data contents (e.g., values), structural elements (e.g., schema relationships), and logs (e.g., query histories). Moreover, the data registry serves as a centralized repository where information such as available indices is stored. This unified interface allows any agent or component to seamless access and utilization of information in a consistent manner.

\subsection{Sessions - Providing Context}

A `session' defines the context and scope for agents' collaborative work. Streams and sessions together enable event-driven orchestration. Each agent signals its entry and exit from the session and creates output streams by posting instructions to the session stream. Sessions can be initiated programmatically or by user action, where they can add agents themselves. For instance, a user might add a \textsc{Profiler} agent to generate job seeker profiles. Additional context can be established by extending the current context, beginning with the session. For instance, multiple interactions aimed at gathering user profile information can be grouped together as \textsc{Session:ID:Profile}, analogous to scoping in programming.

Subsequently, other agents can be included in the session explicitly by the user, via session configuration, or through a task planner. Agents can also behave autonomously by monitoring streams within the session. If they decide to listen, they process data in the input stream and generate data (or instructions) into new output streams within the session. Streams and messages are tagged to enable selective consumption by other agents.

\subsection{Task Planner - Planning Work for Agents}

While the architecture allows agents to operate autonomously using stream and/or message tags, in open-ended scenarios such as conversational interactions, a task planner is essential to guide meaningful discourse. A task planner interprets user requests and devises a task plan that available agents can execute on plan steps and pass on data and instructions to other agents in a workflow manner. To seamlessly integrate into the architecture, we model the task planner as an agent itself. It listens to the initial user stream and formulates a task plan structured as directed acyclic graphs (DAGs) connecting agent input and outputs, and emits the plan into a stream. Each node within these DAGs represents a sub-task assigned to a specific agent. The task planner utilizes metadata sourced from the agent registry to identify suitable agents for each sub-task.

\begin{figure}[!htb] 
  \centering
  \includegraphics[width=\linewidth]{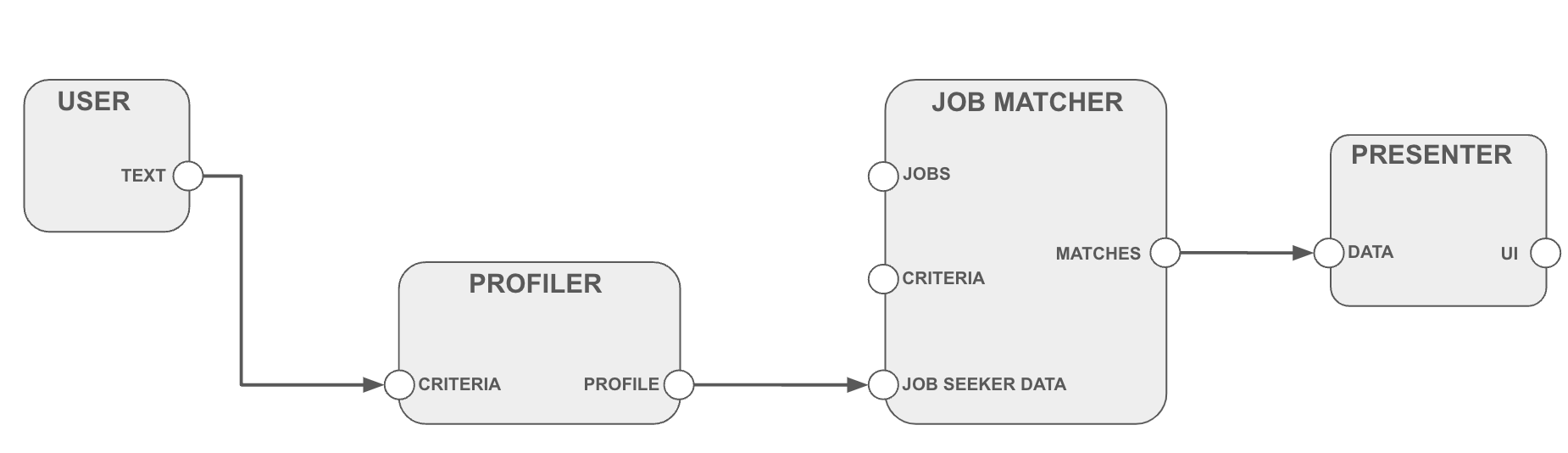}
  \caption{A Task Plan Example: Connecting input and output parameters of agents.}
  \label{fig:task-planner} 
  \vspace{-10pt}
\end{figure}

In our example, for the user request ``\textit{I am looking for a data scientist position in sf bay area.}'', the task planner might propose a sequence of actions: first, gather additional background information from the job seeker; next, match the job seeker's profile with available job listings; finally, present the matched jobs to the end-user. These actions can be mapped to available agents such as \textsc{Profiler}, \textsc{Job Matcher}, and \textsc{Presenter} with  input and output parameters matched (Figure \ref{fig:task-planner}). Once this DAG is formulated, similar to other agents, the task planner outputs the plan to a stream to be executed.

The task planner can be interactive, initially presenting a plan to the user, in text form or as a UI, facilitating collaborative planning. The plan can also be dynamic and incremental, meaning it evolves step by step rather than being predetermined in its entirety. Additionally, the task planner can be adaptive, learning from feedback. This feedback could come directly from the user on a per-plan basis or could leverage historical data from previous interactions.

\subsection{Data Planner - Planning queries, Optimizing} 

Data planner's job is to provide agents with the right data. This can manifest in several ways, including (1) agents themselves invoking data planner (using APIs) to find and query data sources  (e.g. \textsc{Profiler} agent might tap a database to formulate relevant questions for the title) (2) task coordinator invoking data planner to transform data from one agent to another (e.g. \textsc{Profiler.Criteria} $\leftarrow$ \textsc{User.Text}).

To support these, data planner needs to interpret data retrieval and transformation task, within the context of task, and decompose it into sub-tasks (e.g. discover, select, join, query, extract, summarize, etc.) to create a plan. For example, \textsc{Profiler.Criteria} $\leftarrow$ \textsc{User.Text}, while at a high-level matches to an `extract' operator, details such as which model to use, along with its configuration, still needs to be figured in the context of the agents (\textsc{Profiler}) and input parameter in question (\textsc{Criteria}) using the metadata in the registry, available models, etc.

Let's consider a more complicated example, e.g. ``\textit{data scientist position in sf bay area.}''. One approach could be to directly employ NL2Q approaches along with data discovery to find the suitable databases and tables and eventually write the query. This approach may not always work, as required data may be distributed across data sources with different modalities. For example, ``\textit{SF bay area}'' won't match any city in the database as such it would required LLM (as a data source) to first create a list of cities in bay area. Similarly,  ``\textit{data scientist}'' might miss some relevant titles, and could require a graph database, which contains a title taxonomy. 

\begin{figure}[!htb] 
  \centering
  \includegraphics[width=\linewidth]{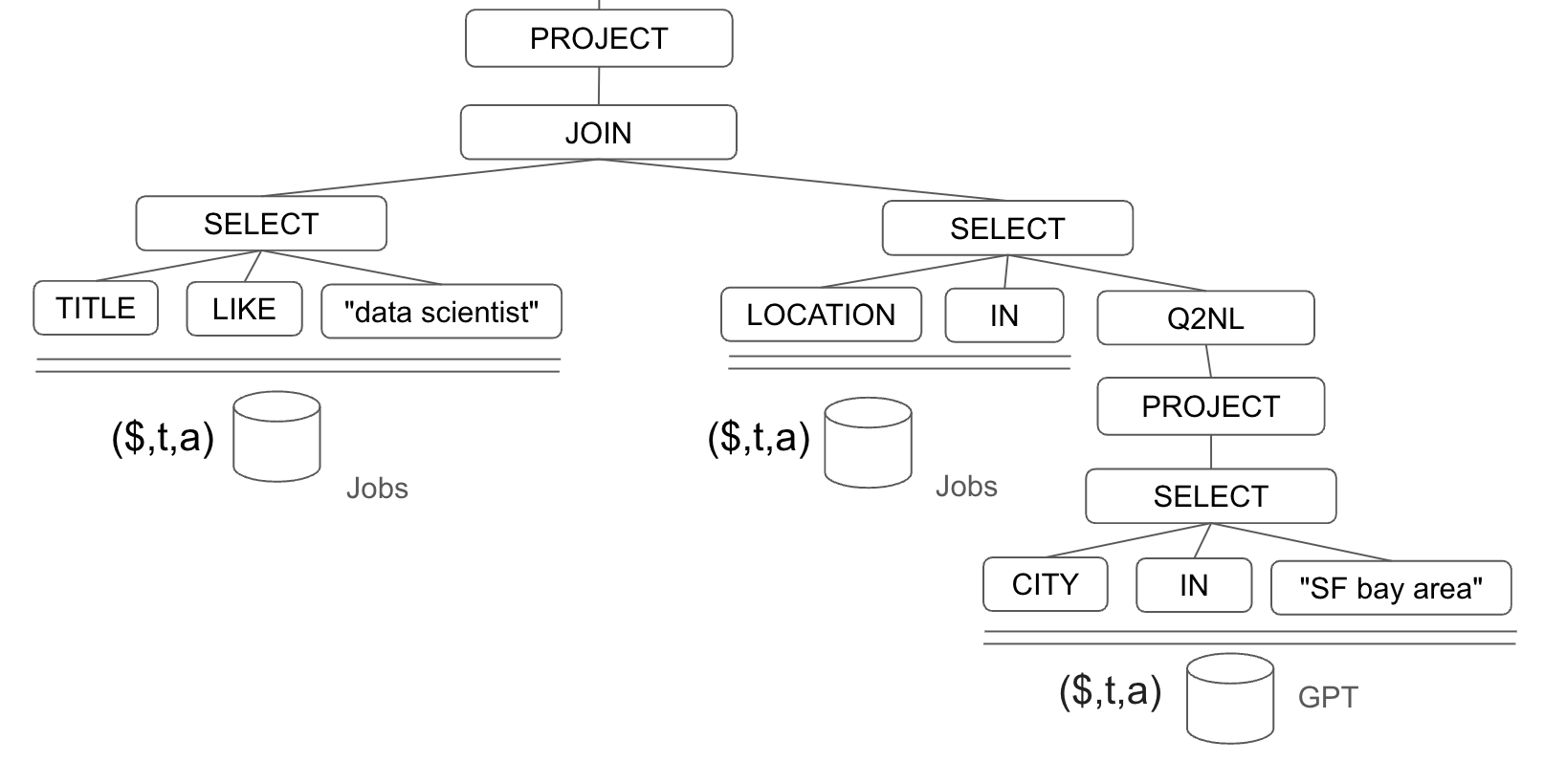}
  \caption{A data plan utilizing a relational table, JOBS, in conjunction with an LLM (GPT), as a data source}
  \label{fig:dataplanner} 
\end{figure}

A decomposition is necessary to break down a query into sub-tasks. This process involves identifying relevant sources for each sub-task using metadata from the data registry and considering the specific requirements of the task. This approach aims to produce a more suitable plan (Figure \ref{fig:dataplanner}). For instance, the query can be broken down into locating cities in the ``\textit{SF bay area}'', identifying titles such as ``\textit{data scientist}'', and then apply a select operator. The data planner needs to identify the right source for each part, for example ``\textit{cities in the SF bay area}'' might be obtained from an OpenAI model, tapping into general knowledge of LLMs. To make this happen the planner needs to inject a new operator \textsc{Q2NL} to transform the part of the query into NL and then figure out details of the operator, such as model specification, prompts, etc. Note also several new operators, beyond established relational operators, need to be introduced to extend data planner capabilities to discover data, handle text operations, etc.

Optimization plays a crucial role in developing suitable data plans especially in production context, as the data planner must also optimize operations within defined constraints such as cost, performance, and quality. This involves optimizing both the overall plan, ensuring it is data-aware during breakdowns, and optimizing individual operators within the plan. For example, this would involve selecting appropriate sources based on metadata (like available indices, model performance, and cost) and fine-tuning individual operators (such as model configurations and prompts) to achieve optimal results.

\subsection{Task Coordinator - Executing Work}

The task planner is concerned with interpreting tasks, while the task coordinator handles execution. The coordinator receives a plan in the form of a DAG, along with an initial budget and projected costs (estimated by the optimizer).  It then initiates agents by streaming instruction messages to them, monitoring the execution of the DAG, and updating the budget with actual costs incurred as the execution progresses. 

The task coordinator directs the execution according to the plan DAG.  It manages the flow of outputs from various agents, ensuring they are correctly utilized as input parameters for subsequent agents. This involves applying any necessary transformations or mappings to ensure compatibility between outputs and inputs as defined by the DAG. For example, one possible sequence of agents and parameters is \textsc{Execute} $\quad$ \textsc{Profiler:} $\quad$ \textsc{Criteria} $\leftarrow $ \textsc{User.Text}. However, in this case \textsc{User.Text} ``\textit{I am looking for a data scientist position in SF bay area.}'' needs to be processed to extract a criteria, e.g. ``\textit{data scientist position in SF bay area.}''. In order to achieve this, the  coordinator invokes the data planner to identify and generate a sequence of data operations (data plan) to  transform output data suitable for input  parameter of the agent. 

The task coordinator monitors the available budget throughout the execution process, tracking factors such as execution time, result quality (e.g. accuracy), and operational costs. If any of these metrics exceed predefined thresholds, the task coordinator may abort the current plan. In such cases, it could potentially trigger the task planner to replan if necessary. Alternatively, it might prompt the user to confirm budget violations before proceeding further.

%% file: casestudy.tex
\section{Case Study: Agentic Employer}
\label{sec:agenticemployer}

\begin{figure}[!htb] 
  \centering
  \includegraphics[width=0.9\linewidth]{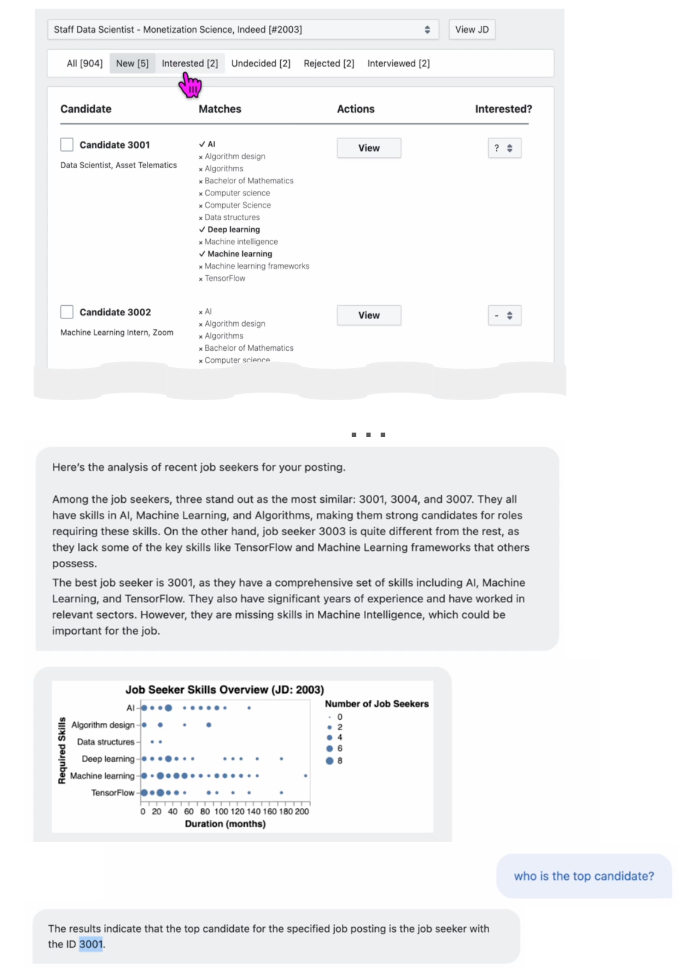}
  \caption{A conversation in Agentic Employer}
  \label{fig:agentic_employer_conversation} 
\end{figure}

Following on the scenario from \ref{sec:example_agentic_employer}, we implemented the blueprint architecture to validate and showcase our proposal. Figure \ref{fig:agentic_employer_conversation} shows a conversation from the implementation where the users (employers) can interact with the applicants through a graphical UI as well as entering text into the conversation. As a results of  these interactions the system utilizing various agents responds by rendering output text and visualizations, and user interface forms that they can interact with. Full details of the use-case application is beyond the scope of this paper, as such we will focus on a few workflows in the application to illustrate how the architecture helped support developing apps with agentic workflows.

The main driver of the application logic is an Agentic Employer agent, which is the first receiver of any user interaction, whether it came in the form of events from the UI/forms, or through text entered into the conversation. 

\begin{figure}[!htb] 
  \centering
  \includegraphics[width=0.55\linewidth]{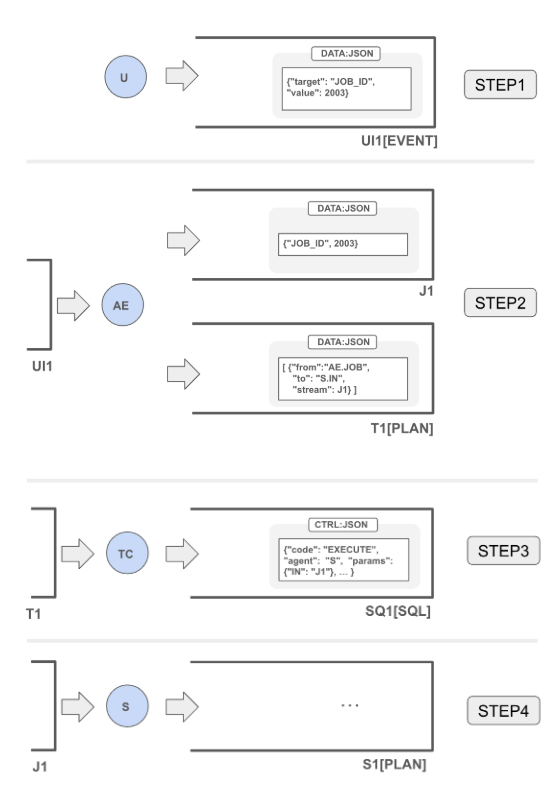}
  \caption{Flow initiated from UI}
  \label{fig:agentic_employer_conversation_ui_flow} 
\end{figure}

Figure \ref{fig:agentic_employer_conversation_ui_flow} shows a sample flow from initiated from the UI interaction. Events from UI are processed just like any other input through streams. When a form is created, an accompanying event stream is generated that the Agentic Employer agents listens to. In Step 1, user agent (U) clicks on the UI to select a job id, which emits an event object into the stream. Upon receiving that Agentic Employer agent (AE) emits the job id into a stream, and creates a plan to invoke a Summarizer agent (S) in Step 2. Task Coordinator agent (TC) listening to any stream with a plan unrolls the plan and emits a Control Message to execute Summarizer agent with the given input (Step 3). Finally, listening to instructions from TC, summarizer invokes its plan to generate a summary.

\begin{figure}[!htb] 
  \centering
  \includegraphics[width=0.55\linewidth]{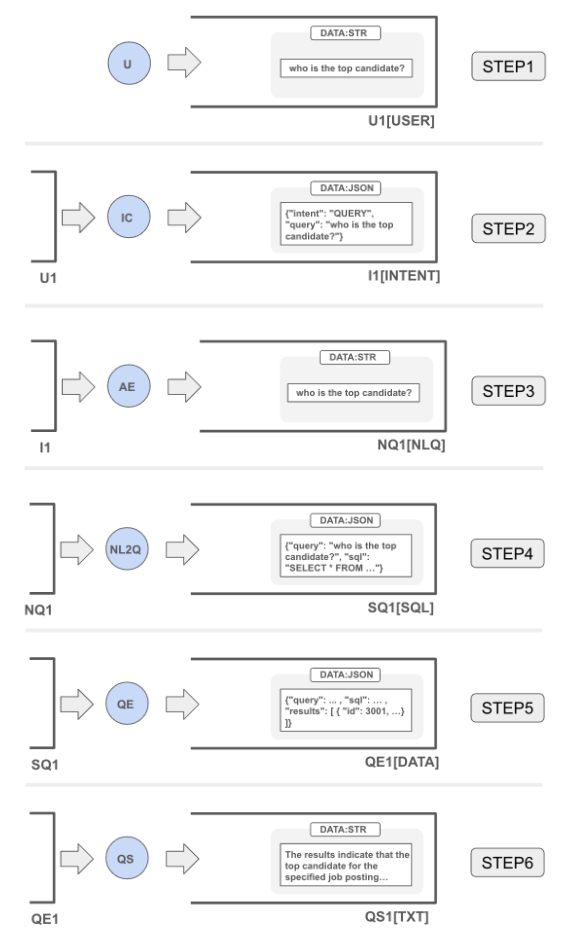}
  \caption{Flow initiated from conversation}
  \label{fig:agentic_employer_conversation_conv_flow} 
\end{figure}

Figure \ref{fig:agentic_employer_conversation_conv_flow} shows a flow initiated by the user entering some text into the conversation, emitted into a stream (Step 1). In the way this application is configured, an Intent Classifier agent (IC) automatically responds by emiting identified intent into the stream. In this case, IC identifies the intent to be an open-ended query (Step 2). Agentic Employer agent then emits the query into a new stream and tags its as NLQ, which is picked up by the NL2Q agent to identify a suitable database query, in this case SQL (Step 3). The next few steps, automatically execute one after another through configuration of the stream tags, where the SQL agent (QE) executes the query from NLQ output, Query Summarizer agent (QS), utilizing LLMs, explains the query results.

%% file: discussion.tex
\section{Discussion}

Our proposal puts forward a blueprint distributed systems architecture for compound AI systems, specifically for enterprise to drive multiple use-cases. While much of the proposal focused on the key abstractions,  components, and flow of data and control with a focus on flexibility and extensibility, much remains to be designed, specifically the key components such as agents, registries, planners, and optimizers themselves. There are numerous intriguing research questions still unanswered in the fields of NLP, AI, databases, HCI, offering ample opportunities for interdisciplinary collaboration. We will discuss some here.

\textbf{Agents.} Effectively and efficiently designed agents are crucial to the successful completion of tasks. In these regards, there are many research questions: How can agents be designed to efficiently adapt to and perform new tasks? How can agents learn and optimize their performance based on feedback and evolving tasks? How to compute a learned representation of an agent based on historical data and performance? How to design effective interactive agents? How to generate user interaction components and handle events that properly interfaces with the architecture?

\textbf{Data and Operators.}
Data serves as the foundation for enterprises, and leveraging it effectively with LLMs is crucial. This opens up new questions: 
How to learn effective representations of data at various granularity, exploiting data hierarchy, queries, modality, and beyond? How to map, connect, and learn cross-modal relationships and representations? How to maintain the data catalog with evolving data and business requirements based on feedback? How to effectively design new benchmarks, metrics, and develop formal problem definitions? What is the universal set of data operators for multiple modalities, including text and graphs? How to design text operators to extract, compare, aggregate, and summarize? How to ensure data-governance and support data privacy in agentic workflows, with potentially agents with different privileges?

\textbf{Planning.} Despite emerging attempts to explore LLMs' reasoning capabilities and use them as planners~\cite{zhao2024large, huang2022language} or `routers' of existing tools and APIs~\cite{liang2024taskmatrix, qin2023toolllm,kim2023llmcompiler}, LLMs alone still cannot solve the planning problem~\cite{kambhampati2024llms,valmeekam2023planning}. Key questions remain, including: How to exploit LLMs for planning, yet add verification and constraints? How to perform planning over multi-modal (relational, graph, documents, parametric) data sources?
How to interact with the user in regards to planning, present and refine plans collaboratively? How to learn feedback and attribute back to agents and operators?

\textbf{Optimization.} Optimization is critical both for task and data planning for production settings, both as a driver of QoS and also business-wise, as cost and performance affects the bottom-line. Optimization is a well-studied subject but new questions emerge: How to perform  cost estimation for (new) operators, given dependence on data (size and beyond)? How to handle uncertainty in sources such as LLMs? How to estimate the overall plan cost? How to incorporate accrued budget into planners?

\textbf{Reliability.} While our proposal is built on distributed systems research (e.g. containers, stream, federated systems, message queues) and applied in practice to scale and operate in a fault-tolerant manner, the `agentic workflows' brings up its own challenges. Unlike deterministic workflows agents, being nondeterministic in nature,  introduces concerns about fault-tolerance and reasoning capacity especially in workflows when chained together \cite{RahmanRetrievalAG, pezeshkpour2024reasoning} and opportunities for research: How to build in fault-tolerance into architecture and agent design to improve reliability? How to design planners with appropriate error handling and re-planning? How to design user experience to involve humans in the analysis, debugging, and mitigation of system/agent error?

%% file: conclusion.tex
\section{Conclusion}
\label{sec:conclusion}

In this paper, we propose a `blueprint' architecture aimed at maximizing the benefits of LLMs while addressing concerns around LLMs. We advocated for a systems approach  to develop dependable, efficient, and user-friendly AI that supports diverse enterprise applications within a unified architecture. While we put forward the abstractions and mechanisms for seamless integration, orchestration, coordination, and optimization, there is still much to be designed, developed, and researched regarding the components themselves.

%% file: main.bbl
\begin{thebibliography}{10}

\bibitem{asai2023self}
Akari Asai, Zeqiu Wu, Yizhong Wang, Avirup Sil, and Hannaneh Hajishirzi.
\newblock Self-rag: Learning to retrieve, generate, and critique through self-reflection.
\newblock {\em arXiv preprint arXiv:2310.11511}, 2023.

\bibitem{cai2023large}
Tianle Cai, Xuezhi Wang, Tengyu Ma, Xinyun Chen, and Denny Zhou.
\newblock Large language models as tool makers.
\newblock {\em arXiv preprint arXiv:2305.17126}, 2023.

\bibitem{chen2023autoagents}
Guangyao Chen, Siwei Dong, Yu~Shu, Ge~Zhang, Jaward Sesay, B{\"o}rje~F Karlsson, Jie Fu, and Yemin Shi.
\newblock Autoagents: A framework for automatic agent generation.
\newblock {\em arXiv preprint arXiv:2309.17288}, 2023.

\bibitem{chen2023frugalgpt}
Lingjiao Chen, Matei Zaharia, and James Zou.
\newblock Frugalgpt: How to use large language models while reducing cost and improving performance.
\newblock {\em arXiv preprint arXiv:2305.05176}, 2023.

\bibitem{chen2023agentverse}
Weize Chen, Yusheng Su, Jingwei Zuo, Cheng Yang, Chenfei Yuan, Chen Qian, Chi-Min Chan, Yujia Qin, Yaxi Lu, Ruobing Xie, et~al.
\newblock Agentverse: Facilitating multi-agent collaboration and exploring emergent behaviors in agents.
\newblock {\em arXiv preprint arXiv:2308.10848}, 2023.

\bibitem{chen2023symphony}
Zui Chen, Zihui Gu, Lei Cao, Ju~Fan, Samuel Madden, and Nan Tang.
\newblock Symphony: Towards natural language query answering over multi-modal data lakes.
\newblock In {\em CIDR}, 2023.

\bibitem{crawford2024bmw}
Noel Crawford, Edward~B Duffy, Iman Evazzade, Torsten Foehr, Gregory Robbins, Debbrata~Kumar Saha, Jiya Varma, and Marcin Ziolkowski.
\newblock Bmw agents--a framework for task automation through multi-agent collaboration.
\newblock {\em arXiv preprint arXiv:2406.20041}, 2024.

\bibitem{du2022glam}
Nan Du, Yanping Huang, Andrew~M Dai, Simon Tong, Dmitry Lepikhin, Yuanzhong Xu, Maxim Krikun, Yanqi Zhou, Adams~Wei Yu, Orhan Firat, et~al.
\newblock Glam: Efficient scaling of language models with mixture-of-experts.
\newblock In {\em International Conference on Machine Learning}, pages 5547--5569. PMLR, 2022.

\bibitem{cmdbench}
Yanlin Feng, Sajjadur Rahman, Aaron Feng, Vincent Chen, and Eser Kandogan.
\newblock Cmdbench: {A} benchmark for coarse-to-fine multimodal data discovery in compound {AI} systems.
\newblock In {\em Proceedings of the Conference on Governance, Understanding and Integration of Data for Effective and Responsible AI, {GUIDE-AI} 2024, Santiago, Chile, June 9-15, 2024}, pages 16--25. {ACM}, 2024.

\bibitem{hao2024toolkengpt}
Shibo Hao, Tianyang Liu, Zhen Wang, and Zhiting Hu.
\newblock Toolkengpt: Augmenting frozen language models with massive tools via tool embeddings.
\newblock {\em Advances in neural information processing systems}, 36, 2024.

\bibitem{huang2022language}
Wenlong Huang, Pieter Abbeel, Deepak Pathak, and Igor Mordatch.
\newblock Language models as zero-shot planners: Extracting actionable knowledge for embodied agents.
\newblock In {\em International conference on machine learning}, pages 9118--9147. PMLR, 2022.

\bibitem{ji2023survey}
Ziwei Ji, Nayeon Lee, Rita Frieske, Tiezheng Yu, Dan Su, Yan Xu, Etsuko Ishii, Ye~Jin Bang, Andrea Madotto, and Pascale Fung.
\newblock Survey of hallucination in natural language generation.
\newblock {\em ACM Computing Surveys}, 55(12):1--38, 2023.

\bibitem{kambhampati2024llms}
Subbarao Kambhampati, Karthik Valmeekam, Lin Guan, Kaya Stechly, Mudit Verma, Siddhant Bhambri, Lucas Saldyt, and Anil Murthy.
\newblock Llms can't plan, but can help planning in llm-modulo frameworks.
\newblock {\em arXiv preprint arXiv:2402.01817}, 2024.

\bibitem{kandogan2024blueprint}
Eser Kandogan, Sajjadur Rahman, Nikita Bhutani, Dan Zhang, Rafael~Li Chen, Kushan Mitra, Sairam Gurajada, Pouya Pezeshkpour, Hayate Iso, Yanlin Feng, et~al.
\newblock A blueprint architecture of compound ai systems for enterprise.
\newblock {\em arXiv preprint arXiv:2406.00584}, 2024.

\bibitem{kasneci2023chatgpt}
Enkelejda Kasneci, Kathrin Se{\ss}ler, Stefan K{\"u}chemann, Maria Bannert, Daryna Dementieva, Frank Fischer, Urs Gasser, Georg Groh, Stephan G{\"u}nnemann, Eyke H{\"u}llermeier, et~al.
\newblock Chatgpt for good? on opportunities and challenges of large language models for education.
\newblock {\em Learning and individual differences}, 103:102274, 2023.

\bibitem{khattab2023dspy}
Omar Khattab, , et~al.
\newblock Dspy: Compiling declarative language model calls into self-improving pipelines.
\newblock {\em arXiv preprint arXiv:2310.03714}, 2023.

\bibitem{kim2023llmcompiler}
Sehoon Kim, Suhong Moon, Ryan Tabrizi, Nicholas Lee, Michael Mahoney, Kurt Keutzer, and Amir Gholami.
\newblock An llm compiler for parallel function calling.
\newblock {\em arXiv}, 2023.

\bibitem{lewis2020retrieval}
Patrick Lewis, Ethan Perez, Aleksandra Piktus, Fabio Petroni, Vladimir Karpukhin, Naman Goyal, Heinrich K{\"u}ttler, Mike Lewis, Wen-tau Yih, Tim Rockt{\"a}schel, et~al.
\newblock Retrieval-augmented generation for knowledge-intensive nlp tasks.
\newblock {\em Advances in Neural Information Processing Systems}, 33:9459--9474, 2020.

\bibitem{lewkowycz2022solving}
Aitor Lewkowycz, Anders Andreassen, David Dohan, Ethan Dyer, Henryk Michalewski, Vinay Ramasesh, Ambrose Slone, Cem Anil, Imanol Schlag, Theo Gutman-Solo, et~al.
\newblock Solving quantitative reasoning problems with language models.
\newblock {\em Advances in Neural Information Processing Systems}, 35:3843--3857, 2022.

\bibitem{li2023self}
Miaoran Li, Baolin Peng, Michel Galley, Jianfeng Gao, and Zhu Zhang.
\newblock Self-checker: Plug-and-play modules for fact-checking with large language models.
\newblock {\em arXiv preprint arXiv:2305.14623}, 2023.

\bibitem{li2023survey}
Yingji Li, Mengnan Du, Rui Song, Xin Wang, and Ying Wang.
\newblock A survey on fairness in large language models.
\newblock {\em arXiv preprint arXiv:2308.10149}, 2023.

\bibitem{liang2021towards}
Paul~Pu Liang, Chiyu Wu, Louis-Philippe Morency, and Ruslan Salakhutdinov.
\newblock Towards understanding and mitigating social biases in language models.
\newblock In {\em International Conference on Machine Learning}, pages 6565--6576. PMLR, 2021.

\bibitem{liang2024taskmatrix}
Yaobo Liang, Chenfei Wu, Ting Song, Wenshan Wu, Yan Xia, Yu~Liu, Yang Ou, Shuai Lu, Lei Ji, Shaoguang Mao, et~al.
\newblock Taskmatrix. ai: Completing tasks by connecting foundation models with millions of apis.
\newblock {\em Intelligent Computing}, 3:0063, 2024.

\bibitem{liu2024declarative}
Chunwei Liu, Matthew Russo, Michael Cafarella, Lei Cao, Peter~Baille Chen, Zui Chen, Michael Franklin, Tim Kraska, Samuel Madden, and Gerardo Vitagliano.
\newblock A declarative system for optimizing ai workloads.
\newblock {\em arXiv preprint arXiv:2405.14696}, 2024.

\bibitem{Liu_LlamaIndex_2022}
Jerry Liu.
\newblock {LlamaIndex}, 11 2022.

\bibitem{liu2024agentlite}
Zhiwei Liu, Weiran Yao, Jianguo Zhang, Liangwei Yang, Zuxin Liu, Juntao Tan, Prafulla~K Choubey, Tian Lan, Jason Wu, Huan Wang, et~al.
\newblock Agentlite: A lightweight library for building and advancing task-oriented llm agent system.
\newblock {\em arXiv preprint arXiv:2402.15538}, 2024.

\bibitem{lu2024chameleon}
Pan Lu, Baolin Peng, Hao Cheng, Michel Galley, Kai-Wei Chang, Ying~Nian Wu, Song-Chun Zhu, and Jianfeng Gao.
\newblock Chameleon: Plug-and-play compositional reasoning with large language models.
\newblock {\em Advances in Neural Information Processing Systems}, 36, 2024.

\bibitem{lu2021fantastically}
Yao Lu, Max Bartolo, Alastair Moore, Sebastian Riedel, and Pontus Stenetorp.
\newblock Fantastically ordered prompts and where to find them: Overcoming few-shot prompt order sensitivity.
\newblock {\em arXiv preprint arXiv:2104.08786}, 2021.

\bibitem{maekawa2024retrieval}
Seiji Maekawa, Hayate Iso, Sairam Gurajada, and Nikita Bhutani.
\newblock Retrieval helps or hurts? a deeper dive into the efficacy of retrieval augmentation to language models.
\newblock In {\em Proceedings of the 2024 Conference of the North American Chapter of the Association for Computational Linguistics: Human Language Technologies (Volume 1: Long Papers)}, pages 5506--5521, 2024.

\bibitem{mishra2023characterizing}
Aditi Mishra et~al.
\newblock Characterizing large language models as rationalizers of knowledge-intensive tasks.
\newblock {\em arXiv:2311.05085}, 2023.

\bibitem{paranjape2023art}
Bhargavi Paranjape, Scott Lundberg, Sameer Singh, Hannaneh Hajishirzi, Luke Zettlemoyer, and Marco~Tulio Ribeiro.
\newblock Art: Automatic multi-step reasoning and tool-use for large language models.
\newblock {\em arXiv preprint arXiv:2303.09014}, 2023.

\bibitem{10.1145/3586183.3606763}
Joon~Sung Park, Joseph O'Brien, Carrie~Jun Cai, Meredith~Ringel Morris, Percy Liang, and Michael~S. Bernstein.
\newblock Generative agents: Interactive simulacra of human behavior.
\newblock In {\em Proceedings of the 36th Annual ACM Symposium on User Interface Software and Technology}, UIST '23, New York, NY, USA, 2023. Association for Computing Machinery.

\bibitem{peterson1977petri}
James~L Peterson.
\newblock Petri nets.
\newblock {\em ACM Computing Surveys (CSUR)}, 9(3):223--252, 1977.

\bibitem{pezeshkpour2024reasoning}
Pouya Pezeshkpour, Eser Kandogan, Nikita Bhutani, Sajjadur Rahman, Tom Mitchell, and Estevam Hruschka.
\newblock Reasoning capacity in multi-agent systems: Limitations, challenges and human-centered solutions.
\newblock {\em arXiv preprint arXiv:2402.01108}, 2024.

\bibitem{qin2023toolllm}
Yujia Qin, Shihao Liang, Yining Ye, Kunlun Zhu, Lan Yan, Yaxi Lu, Yankai Lin, Xin Cong, Xiangru Tang, Bill Qian, et~al.
\newblock Toolllm: Facilitating large language models to master 16000+ real-world apis.
\newblock {\em arXiv preprint arXiv:2307.16789}, 2023.

\bibitem{RahmanRetrievalAG}
Sajjadur Rahman, Dan Zhang, Nikita Bhutani, Estevam~R. Hruschka, and Eser Kandogan.
\newblock Retrieval augmented generation in the wild: A system 2 perspective.
\newblock {\em IEEE Data Engineering Bulletin}, 48(4):47--70, 2024.

\bibitem{ram2023context}
Ori Ram, Yoav Levine, Itay Dalmedigos, Dor Muhlgay, Amnon Shashua, Kevin Leyton-Brown, and Yoav Shoham.
\newblock In-context retrieval-augmented language models.
\newblock {\em Transactions of the Association for Computational Linguistics}, 11:1316--1331, 2023.

\bibitem{schick2024toolformer}
Timo Schick, Jane Dwivedi-Yu, Roberto Dess{\`\i}, Roberta Raileanu, Maria Lomeli, Eric Hambro, Luke Zettlemoyer, Nicola Cancedda, and Thomas Scialom.
\newblock Toolformer: Language models can teach themselves to use tools.
\newblock {\em Advances in Neural Information Processing Systems}, 36, 2024.

\bibitem{semnani2023wikichat}
Sina~J Semnani, Violet~Z Yao, Heidi~C Zhang, and Monica~S Lam.
\newblock Wikichat: Stopping the hallucination of large language model chatbots by few-shot grounding on wikipedia.
\newblock {\em arXiv preprint arXiv:2305.14292}, 2023.

\bibitem{thoppilan2022lamda}
Romal Thoppilan, Daniel De~Freitas, Jamie Hall, Noam Shazeer, Apoorv Kulshreshtha, Heng-Tze Cheng, Alicia Jin, Taylor Bos, Leslie Baker, Yu~Du, et~al.
\newblock Lamda: Language models for dialog applications.
\newblock {\em arXiv preprint arXiv:2201.08239}, 2022.

\bibitem{topsakal2023creating}
Oguzhan Topsakal and Tahir~Cetin Akinci.
\newblock Creating large language model applications utilizing langchain: A primer on developing llm apps fast.
\newblock In {\em International Conference on Applied Engineering and Natural Sciences}, volume~1, pages 1050--1056, 2023.

\bibitem{valmeekam2023planning}
Karthik Valmeekam, Matthew Marquez, Sarath Sreedharan, and Subbarao Kambhampati.
\newblock On the planning abilities of large language models-a critical investigation.
\newblock {\em Advances in Neural Information Processing Systems}, 36:75993--76005, 2023.

\bibitem{wang2023gemini}
Yuqing Wang and Yun Zhao.
\newblock Gemini in reasoning: Unveiling commonsense in multimodal large language models.
\newblock {\em arXiv preprint arXiv:2312.17661}, 2023.

\bibitem{wang2024tools}
Zhiruo Wang, Zhoujun Cheng, Hao Zhu, Daniel Fried, and Graham Neubig.
\newblock What are tools anyway? a survey from the language model perspective.
\newblock {\em arXiv preprint arXiv:2403.15452}, 2024.

\bibitem{wang2024chain}
Zilong Wang, Hao Zhang, Chun-Liang Li, Julian~Martin Eisenschlos, Vincent Perot, Zifeng Wang, Lesly Miculicich, Yasuhisa Fujii, Jingbo Shang, Chen-Yu Lee, et~al.
\newblock Chain-of-table: Evolving tables in the reasoning chain for table understanding.
\newblock {\em arXiv preprint arXiv:2401.04398}, 2024.

\bibitem{wei2022chain}
Jason Wei, Xuezhi Wang, Dale Schuurmans, Maarten Bosma, Fei Xia, Ed~Chi, Quoc~V Le, Denny Zhou, et~al.
\newblock Chain-of-thought prompting elicits reasoning in large language models.
\newblock {\em Advances in neural information processing systems}, 35:24824--24837, 2022.

\bibitem{wu2024exploring}
Likang Wu, Zhaopeng Qiu, Zhi Zheng, Hengshu Zhu, and Enhong Chen.
\newblock Exploring large language model for graph data understanding in online job recommendations.
\newblock In {\em Proceedings of the AAAI Conference on Artificial Intelligence}, volume~38, pages 9178--9186, 2024.

\bibitem{wu2023autogen}
Qingyun Wu, Gagan Bansal, Jieyu Zhang, Yiran Wu, Shaokun Zhang, Erkang Zhu, Beibin Li, Li~Jiang, Xiaoyun Zhang, and Chi Wang.
\newblock Autogen: Enabling next-gen llm applications via multi-agent conversation framework.
\newblock {\em arXiv preprint arXiv:2308.08155}, 2023.

\bibitem{wu2309hayate}
Yunshu Wu.
\newblock Hayate iso, pouya pezeshkpour, nikita bhutani, and estevam hruschka. 2023b. less is more for long document summary evaluation by llms.
\newblock {\em arXiv preprint arXiv:2309.07382}.

\bibitem{xi2023rise}
Zhiheng Xi, Wenxiang Chen, Xin Guo, Wei He, Yiwen Ding, Boyang Hong, Ming Zhang, Junzhe Wang, Senjie Jin, Enyu Zhou, et~al.
\newblock The rise and potential of large language model based agents: A survey.
\newblock {\em arXiv preprint arXiv:2309.07864}, 2023.

\bibitem{Xie2023openagents}
Tianbao Xie, Fan Zhou, Zhoujun Cheng, Peng Shi, Luoxuan Weng, Yitao Liu, Toh~Jing Hua, Junning Zhao, Qian Liu, Che Liu, et~al.
\newblock Openagents: An open platform for language agents in the wild.
\newblock {\em arXiv preprint arXiv:2310.10634}, 2023.

\bibitem{yao2022react}
Shunyu Yao, Jeffrey Zhao, Dian Yu, Nan Du, Izhak Shafran, Karthik Narasimhan, and Yuan Cao.
\newblock React: Synergizing reasoning and acting in language models.
\newblock {\em arXiv preprint arXiv:2210.03629}, 2022.

\bibitem{yoran2023making}
Ori Yoran, Tomer Wolfson, Ori Ram, and Jonathan Berant.
\newblock Making retrieval-augmented language models robust to irrelevant context.
\newblock {\em arXiv preprint arXiv:2310.01558}, 2023.

\bibitem{yuan2024rigorllm}
Zhuowen Yuan, Zidi Xiong, Yi~Zeng, Ning Yu, Ruoxi Jia, Dawn Song, and Bo~Li.
\newblock Rigorllm: Resilient guardrails for large language models against undesired content.
\newblock {\em arXiv preprint arXiv:2403.13031}, 2024.

\bibitem{compound-ai-blog}
Matei Zaharia, Omar Khattab, Lingjiao Chen, Jared~Quincy Davis, Heather Miller, Chris Potts, James Zou, Michael Carbin, Jonathan Frankle, Naveen Rao, and Ali Ghodsi.
\newblock The shift from models to compound ai systems.
\newblock \url{https://bair.berkeley.edu/blog/2024/02/18/compound-ai-systems/}, 2024.

\bibitem{zhang2023ecoassistant}
Jieyu Zhang, Ranjay Krishna, Ahmed~H Awadallah, and Chi Wang.
\newblock Ecoassistant: Using llm assistant more affordably and accurately.
\newblock {\em arXiv preprint arXiv:2310.03046}, 2023.

\bibitem{zhao2023verify}
Ruochen Zhao, Xingxuan Li, Shafiq Joty, Chengwei Qin, and Lidong Bing.
\newblock Verify-and-edit: A knowledge-enhanced chain-of-thought framework.
\newblock {\em arXiv preprint arXiv:2305.03268}, 2023.

\bibitem{zhao2024large}
Zirui Zhao, Wee~Sun Lee, and David Hsu.
\newblock Large language models as commonsense knowledge for large-scale task planning.
\newblock {\em Advances in Neural Information Processing Systems}, 36, 2024.

\bibitem{zhou2023agents}
Wangchunshu Zhou, Yuchen~Eleanor Jiang, Long Li, Jialong Wu, Tiannan Wang, Shi Qiu, Jintian Zhang, Jing Chen, Ruipu Wu, Shuai Wang, et~al.
\newblock Agents: An open-source framework for autonomous language agents.
\newblock {\em arXiv preprint arXiv:2309.07870}, 2023.

\bibitem{zhu2024knowagent}
Yuqi Zhu, Shuofei Qiao, Yixin Ou, Shumin Deng, Ningyu Zhang, Shiwei Lyu, Yue Shen, Lei Liang, Jinjie Gu, and Huajun Chen.
\newblock Knowagent: Knowledge-augmented planning for llm-based agents.
\newblock {\em arXiv preprint arXiv:2403.03101}, 2024.

\end{thebibliography}
